\documentclass[conference]{IEEEtran}
\IEEEoverridecommandlockouts
\usepackage[noadjust]{cite}
\usepackage{amsmath,amssymb,amsfonts}
\usepackage{algorithmic}
\usepackage{graphicx}
\usepackage{textcomp}
\usepackage{xcolor}
\usepackage{comment}
\usepackage{multirow} 
\usepackage{footnote}
\usepackage{verbatim}
\usepackage{graphicx}
\usepackage[hyphens]{url}
\usepackage{listings}
\usepackage{amsmath}
\usepackage{hyperref}
\usepackage{pslatex}
\usepackage{amsfonts}

\usepackage{amssymb}
\usepackage{url}
\usepackage{amsmath}
\usepackage[utf8]{inputenc}
\usepackage{tabularx}
\usepackage{adjustbox}
\usepackage{lscape}
\usepackage{booktabs}
\usepackage{graphicx}
\usepackage{tikz}
\usepackage{float}
\usepackage{subcaption}
\usetikzlibrary{shapes,backgrounds,calc,arrows}
\usepackage{amsmath} 
\usepackage{microtype}

\def\BibTeX{{\rm B\kern-.05em{\sc i\kern-.025em b}\kern-.08em
    T\kern-.1667em\lower.7ex\hbox{E}\kern-.125emX}}

\usepackage{times}
\usepackage{latexsym}

\usepackage{placeins}
\usepackage{longtable}
\usepackage{acronym}
\usepackage{stfloats}
\usepackage{adjustbox}
\usepackage{breqn}
\usepackage{amsmath}
\usepackage{url}

\usepackage[T1]{fontenc}

\usepackage[utf8]{inputenc}

\usepackage{microtype}
\usepackage{comment}

\usepackage{soul}
\usepackage{epstopdf}
\usepackage{booktabs}
\usepackage{graphicx}
\begin{document}

\title{ReviewGraph: A Knowledge Graph Embedding Based Framework for Review Rating Prediction with Sentiment Features }

\author{
           A.J.W. de Vink$^{1}$, Natalia Amat-Lefort$^{1}$, Lifeng Han$^{*, 1,2}$
              \vspace*{0.075cm}
\\            $^1$ LIACS, Leiden University
  \\          $^2$ LUMC, Leiden University, NL  
    \\        {\tt bertdevinkheusden@gmail.com n.amat.lefort, l.han@liacs.leidenuniv.nl} 
    %
      \\   $^*$\textit{Corresponding Author} 
}

\maketitle

\begin{abstract}
In the hospitality industry, understanding the factors that drive customer review ratings is critical for improving guest satisfaction and business performance. This work proposes ReviewGraph for Review Rating Prediction (RRP), a novel framework that transforms textual customer reviews into knowledge graphs by extracting (subject, predicate, object) triples and associating sentiment scores. Using graph embeddings (Node2Vec) and sentiment features, the framework predicts review rating scores through machine learning classifiers. We compare ReviewGraph performance with traditional NLP baselines (such as Bag of Words, TF-IDF, and Word2Vec) and large language models (LLMs), evaluating them in the HotelRec dataset. In comparison to the state of the art literature, our proposed model performs similar to their best performing model but with lower computational cost (without ensemble).
 While ReviewGraph achieves comparable predictive performance to LLMs and outperforms baselines on agreement-based metrics such as Cohen's Kappa, it offers additional advantages in interpretability, visual exploration, and potential integration into Retrieval-Augmented Generation (RAG) systems. This work highlights the potential of graph-based representations for enhancing review analytics and lays the groundwork for future research integrating advanced graph neural networks and fine-tuned LLM-based extraction methods. We will share ReviewGraph output and platform open-sourced on our GitHub page \url{https://github.com/aaronlifenghan/ReviewGraph}


\end{abstract}

\section{Introduction} \label{introduction}

In the past, consumers had a limited way to measure the quality of services within the hospitality industry. The main external source of information for them in this regard were professional reviewers. As the internet age went on, however, professional reviewers are increasingly replaced by other consumer reviewers for this purpose \cite{Viglia2016}.
Consumer reviews affect up to 50 percent of booking decisions \cite{Duan2016}. It also affects a multitude of metrics regarding hotel performance, among which are: occupancy rates, revenue per available room, guest satisfaction, and brand reputation. In particular, negative reviews affect these metrics strongly \cite{gabbard2023}. Taking these facts into account, it is of great interest for hotels to be able to accurately estimate what factors affect their review ratings. 
 
We propose a new framework called \textbf{ReviewGraph} \footnote{Peer-reviewed and Published version in ICKG-2025 proceedings \url{https://cyprusconferences.org/ickg2025/} (The 16th IEEE International Conference on Knowledge Graphs, November 13-14, 2025, Limassol, Cyprus).} that will help hoteliers, and business owners generally by converting their reviews into a knowledge graph. We believe that this will allow a more context-aware and granular representation of what reviews say, allow users to inspect what causes low-rated reviews more easily, and in the future could be used for Retrieval Augmented Generation of user reviews to quickly have LLMs answer questions about their reviews.   
To investigate this topic, we have formulated the following research questions:
 \textbf{Research Question 1:} \textit{How can knowledge graph embeddings and sentiment prediction models contribute to customer review based rating predictions?}
  \textbf{Research Question 2:} \textit{How do current LLMs perform on review score prediction tasks, in comparison to task-specific NLP models?}
   



%
In summary, the main \textbf{contributions} of this paper are: I), the design and implementation of a proof of concept of the ReviewGraph framework; II), an empirical evaluation comparing our implemented framework to classical baselines and an LLM baseline; and  III), an analysis of the trade-offs between these baselines and our proposed model.







\section{Background and Related Work}\label{relatedwork}

\subsection{Background}
\subsubsection{Word Embeddings and Language Models}
Traditional Natural Language Processing has mainly relied on embedding techniques such as Bag-of-Words or TF-IDF, which ignored word order or semantic similarity. The introduction of newer embedding techniques such as Word2Vec \cite{mikolov2013efficient} addressed this limitation by representing words as vectors where semantic similarity is reflected by its proximity. 
Word2Vec paved the way for embedding techniques and language models that also consider the order within the sentence dynamically such as ELMo \cite{peters2018deep} and BERT \cite{devlin2019bert}.

\subsubsection{Large Language Models}
More recently, Large Language Models (LLMs) such as GPT-4 \cite{openai2023gpt4} are being used for a wide variety of tasks which have been trained on massive datasets, and transformer-based architectures.   
 Because of their architecture and large dataset, they are increasingly used not just for generating text, but also for extraction tasks. In this study, we will also utilize LLMs to predict review ratings. 
\subsubsection{Classification task in NLP}
Text classification is a common problem in the field of NLP defined as the use of a machine learning model, referred to as a classifier, that is trained and used to differentiate between predefined classes based on features that have been extracted from the text in some manner  \cite{dogra2022complete} or prompting LLMs nowadays \cite{cui2023medtem2}. 
 The accuracy of the task depends on the granularity of the method, and on how clearly separable the different text units are from one another \cite{dogra2022complete}.
\subsubsection{Knowledge Graphs}
Knowledge graphs are structured representations of information in the form of a graph. They use a graph-based data model to capture knowledge at scale using information from many different sources. It allows the maintainers to forego a schema and let the database evolve organically \cite{10.1145/3447772}. Knowledge Graphs have garnered extra attention in recent times because of its potential for improving RAG (Retrieval Augmented Generation) using the GraphRAG approach \cite{edge2025localglobalgraphrag}. 

\subsection{Related Work}

\subsubsection{Customer Review Predictions}
Kumar et al. (2024) present an ensemble learning approach for predicting hotel ratings based on user-generated review text content \cite{Kumar2024}. The review data they used is from the same dataset that this paper used. The research makes use of classical word embedding methods such as Bag of Words, TF-IDF, and Word2Vec, and compares these embedding methods to each other. To predict the ratings they make use of an ensemble of classifier models, among which are: Stochastic Gradient Descent Classifier, Logistic Regression, Support Vector Classifier, Random Forest Classifier, K-Nearest Neighbors, and Decision Tree. For the ensemble method, Kumar et al. used hard voting combined with a majority vote to decide the predicted rating score. The only metric they used is accuracy score, and their final highest accuracy score was 57\% with Bag of Words embedding plus the ensemble classifier model. 
In our investigation, we aimed for this accuracy score, and we used their methods as baselines to compare our results to using our new method. As well as using the same dataset. Additionally, they did not make use of sentiment scores while we do.

\subsubsection{Review Visualisation in Hospitality}
As more users generate online reviews, it becomes increasingly difficult for both consumers and businesses to quickly summarize and understand opinions about products and services. One of the earlier works in this space proposed an interactive visualization system that presents commonly mentioned topics and their associated average sentiments using aspect-based sentiment analysis and topic modeling \cite{Ishizuka2015}.

More recently, Huth et al. introduced ViSCitR (Visual Comparison of Cities through Reviews), a system designed to improve the analysis of review data by providing summaries of both positive and negative opinions for each topic \cite{Huth2024}. ViSCitR combines several coordinated views. This includes a geographic map, numeric rating charts across fixed categories, temporal rating trends, and textual summaries of frequently mentioned points. 
However, ViSCitR also has key limitations. Its design emphasizes only recurring patterns, deliberately discarding less frequent or minority opinions to maintain clarity, and it reduces sentiment to positive/negative counts without modeling the intensity of this senetiment. Additionally, all review aspects must fit into a predefined schema of topics, meaning more niche topics may be generalized into generic topics. For example, the unclean bathroom may be generalized as cleanliness (or bathroom). The tool also does not offer predictive analytics or structural analysis to uncover hidden patterns or influential factors beyond the presented summaries. 

Our proposed ReviewGraph model addresses these limitations by adopting a different approach: representing review content as a semantic knowledge graph built from subject–predicate–object triples extracted from the reviews using either LLM or non-LLM methods. Unlike the aggregation-driven approach of ViSCitR, in ReviewGraph any noun or entity mentioned in the reviews can appear in the graph, eliminating the need for predefined categories. This means the system can show any detailed, unexpected topics mentioned in the reviews. Such as WiFwe speed or balcony views. In addition, it can be generalized to reviews outside of the hospitality sector.

\subsubsection{Triple Extraction for Knowledge Graph Construction}
To be able to generate the embeddings for our model, we first need to construct the knowledge graph from the textual review data. Most research in this space is focused on NER (Named Entity Recognition) and relationship extraction \cite{ZHAO2023225}. As we are not just trying to extract commonly mentioned entities (though that is an important part of it), these methods are not satisfactory. It would not recognise sentences such as "it was great" or "it was terrible", missing important context.  
 
As an alternative we investigated Large Language Model based method such as the one by Zhang et. al. (2024) \cite{zhang-etal-2024-fine-tuning}. This method involves fine-tuning an LLM to extract full "triples" from text data as follows:  ⟨ subject, predicate, object ⟩. These methods, while promising, are computationally intensive and time-consuming. Moreover, our study already investigates the effectiveness of LLMs in review rating prediction, making the additional complexity of fine-tuning less justifiable.

Given these considerations, we opted to use the Stanford CoreNLP Open Information Extraction (OpenIE) method \cite{angeli2015leveraging}. It is a non-LLM method that extracts full triples as ⟨ subject, predicate, object ⟩ similarly to the fine-tuned LLM method from the previous paragraph. 

\subsubsection{Knowledge Graph Usage in Hospitality and Business }
In the hospitality industry most investigations on the application of knowledge graphs has involved improving recommendation systems to better recommend hotels to users \cite{Cadeddu2024Optimizing}.   
In the broader business context, Knowledge Graphs have been attracting increasing attention because of their inherent usefulness to aid with Retrieval Augmented Generation \cite{edge2025localglobalgraphrag}.

\section{Methodology}
\label{methods}
To test our research questions, we will use a dataset of hotel reviews to construct a knowledge graph representing relationships between hotels, review aspects, and sentiment information. From this graph, we will generate node and/or edge embeddings that can be used as input features for machine learning models, particularly classification algorithms aimed at predicting review scores. 
 
The models will be evaluated using multiple metrics: accuracy (to assess exact match performance), mean absolute error (MAE) and root mean squared error (RMSE) (to quantify how far off predictions are from true values), and Cohen’s Kappa (to measure agreement between predicted and actual scores while accounting for chance agreement). The scores of these models will be compared to a baseline using embeddings based on more classical natural language processing methods. 
 
To investigate the second research question and hypotheses we will use an LLM fine-tuned on 200 reviews and their ratings to predict the ratings for all the reviews in our dataset. Its performance will be compared against the other task-specific NLP methods we described earlier. Evaluation will be based on the same metrics: accuracy, MAE, RMSE, and Cohen's Kappa. This will allow us to assess whether LLMs can match or outperform the task-specific models on this review rating prediction task, and at what computational or practical cost.





\subsection{Dataset}
The dataset we used is a fraction of the HotelRec \cite{antognini2020hotelrec} dataset. Which contains roughly 50 million reviews on hotels from TripAdvisor. We used only the first 10,000 reviews in this dataset for this study. Each review in the database contains the following features: 
\begin{itemize}
    \item \textbf{hotel\_url:} Unique identifier for the review, indirectly contains the hotel name. The rest of this feature will be discarded in preprocessing.
    \item \textbf{author:} Name of the author of the review, will be disregarded during preprocessing.
    \item \textbf{date:} Date of when the review was posted, will be disregarded during preprocessing.
    \item \textbf{rating:} The rating that the user gave with this review. This is the value that we want to be able to predict in this study.
    \item \textbf{title:} The title of the review, is in text form.
    \item \textbf{text:} The textual body of the review. This is the main feature we will use to predict the rating.
    \item \textbf{property\_dict:} These are values that TripAdvisor itself extracts from the reviews. It is a set of ratings for the various properties the hotel may have. Among these are 'sleep quality', 'rooms', 'service', and 'cleanliness'. This will also be disregarded during preprocessing.
\end{itemize}
Among these 10,000 reviews were 59 different unique hotels. The average rating of all these reviews is 4.162, with a standard deviation of 1.08. 
The average length of a review is 133 characters, while the median review length is 94. 
Any reviews longer than 1000 words were removed from this chart for readability purposes.
For data split, to test the validity of our results we also make use of \textbf{10-fold cross validation} in addition to our \textbf{80-20\% train-test split}, as it has been deployed in the literature for model optimisation and de-biasing the model testing \cite{moreau2018semantic,maldonado2017detection}. 


\subsection{Baseline Models}
\begin{figure}[t]
    \centering
    \includegraphics[width=1\linewidth]{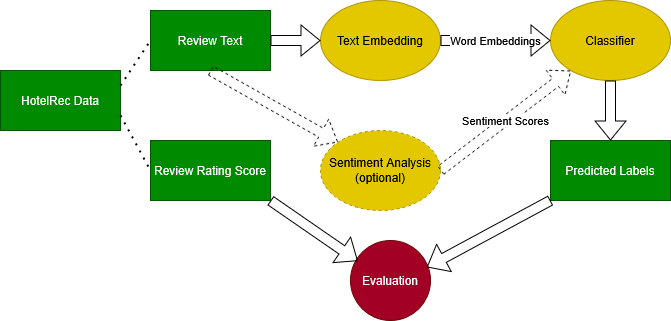}
    \caption{Base Model Design. The green boxes represent the stored input and output data. The yellow ellipses represent processing modules, which can be interchanged with different embedding or classifier models to assess whether performance improves. The red circle represents the evaluation step, where various metrics are used to measure how well the chosen combination of processing modules (in the yellow ellipses) performed. The solid arrows represent the main data flow, the dotted arrow marks an optional sentiment analysis step, and the dotted lines indicate where the input data is split between review text and rating labels.}
    \label{fig:base-model-diagram}
\end{figure}

The baseline model follows a traditional supervised machine learning pipeline for text classification, aiming to predict review rating scores from textual hotel reviews. As shown in Figure ~\ref{fig:base-model-diagram}, the process begins with the HotelRec dataset, which provides both review text and the corresponding rating score.

The review text is processed through a Text Embedding step, where textual data is transformed into numerical vectors using embedding techniques such as either Bag-of-Words, TF-IDF, or Word2Vec. It is yellow because this step is interchangeable with any of these three embedding methods. These word embeddings are then used as input features for a Classifier, which is also interchangeable. This Classifier is then trained to predict the review's rating score. 
 
Optionally, sentiment analysis can be done on the review text to create another feature, which is the sentiment of the review.   
 
The model outputs predicted ratings, which are then compared against the ground-truth Review Rating Scores. This comparison is handled in the evaluation phase, where performance metrics such as accuracy score, MAE, RMSE, and Cohen's Kappa are calculated.

\subsubsection{Preprocessing and Embedding}
Before applying any machine learning models to the review text, we perform a series of preprocessing steps to clean and standardize the input data. These steps are necessary to improve the quality of the features generated by text vectorization techniques such as Bag-of-Words (BoW), Term Frequency–Inverse Document Frequency (TF-IDF), and Word2Vec.

The raw review text is first lowercased and tokenized, which ensures consistency and enables further processing on individual words. We also remove punctuation and special characters, which generally do not contribute meaningful information in this context.

For BoW and TF-IDF, we additionally apply stopword removal and lemmatization. Removing common stopwords (for example, “the”, “is”, “and”) helps reduce noise and the dimensionality of the resulting sparse feature vectors. Lemmatization normalizes words to their base forms,  running for example becomes run, allowing related terms to be grouped together and reducing vocabulary size. These techniques are commonly used in traditional text classification pipelines to improve model efficiency and interpretability.

In contrast, Word2Vec embeddings are sensitive to word context and order. Therefore, we avoid lemmatization or stemming when using Word2Vec, as the model benefits from learning distinctions between different word forms. Word2Vec requires a clean but minimally altered text corpus to effectively learn semantic relationships based on word co-occurrences.

The preprocessing pipeline is implemented using standard tools from the \texttt{nltk} \cite{bird2009natural} and \texttt{scikit-learn} \cite{pedregosa2011scikit} libraries. The result is a set of clean, tokenized text inputs ready for feature extraction using the respective embedding techniques.
\subsubsection{Sentiment Analysis}
Optionally, VADER \cite{hutto2014vader} sentiment analysis can be used to generate a sentiment score for the review, and use this as an additional feature when training the model.
\subsubsection{Model Selection}
For the model we have a selection of classifier models which we use. These are as follows: Random Forest Classifier, Logistic Regression (with Max Entropy Loss), and Multi-Layer Perceptron Classifier. In addition, we use a Dummy Classifier that uses the most common value to compare to.
\subsection{ReviewGraph Model}

\begin{figure}[t]
    \centering
    \includegraphics[width=1\linewidth]{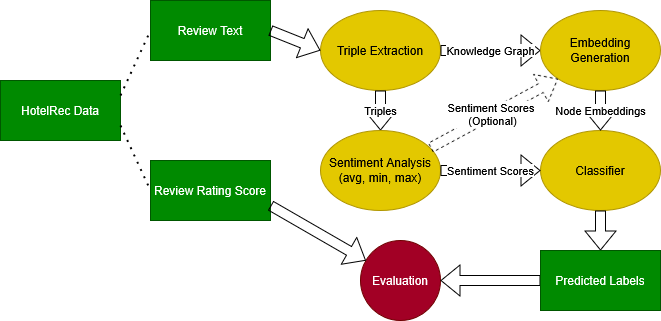}
    \caption{ReviewGraph Model Design: with KG Embedding. 
    }
    \label{fig:enter-label}
\end{figure}

\subsubsection{Preprocessing and Triple Extraction}

For the ReviewGraph model, preprocessing is applied directly to the review text before triple extraction and graph construction. These steps are designed to normalize, clean, and translate the text to ensure consistency and improve the quality of extracted information.

Each review is first translated to English, if necessary, using automatic language detection and the Google Translate APwe \cite{googletrans}. Next, contractions are expanded, for example: “can’t” $ \rightarrow$ “cannot”, to standardize expressions. The text is then cleaned by removing HTML tags, hyphens, apostrophes, and similar punctuation that may interfere with parsing. 

To improve the linguistic structure and ensure consistency in extraction, the script inserts spaces after punctuation marks, a step that aids with sentence segmentation. 
 
After this, the triples are extracted using the StanfordOpenIE library. This returns a .csv file with the subject, predicate, and object of each respective triple. 
 This preprocessing pipeline ensures that reviews, which are often noisy, diverse, and unstructured, are transformed into a consistent format suitable for graph embedding.

\subsubsection{Sentiment Analysis}
Optionally, sentiment analysis can now be performed on the list of triples using the VADER sentiment analyzer \cite{hutto2014vader}. This returns a sentiment score that can range from -1 to 1. This score is stored in the same csv file, in a separate column.
\subsubsection{Graph Embedding}
Once the triples are extracted from the model and optionally have their sentiment values calculated, we use these triples to construct the knowledge graph. Every triple which contains either a node or relationship of 14 characters or longer is discarded during this process. Both the subject, object, and predicate are "normalized", by having their capital letters, punctuation, and other special symbols removed. Additionally, spaces are replaced with underscore symbols. For each word there is also lemmatization and synonym mapping performed. This assures that the amount of different nodes is minimized, to have a more effective model with less noise. 
 
Once that is done, for each object and subject a node is created with a relationship created based on the predicate. Each of these nodes is connected to a node representing the review. The review node in turn is connected to a node representing the hotel. If the object/subject is in the list of amenities, it instead gets a special node type called "amenity", otherwise its node type is called "word". This is only relevant for the system visualization. 
 Once this graph has been constructed using NetworkX \cite{hagberg2008exploring}. It is then converted into a Neo4J \cite{neo4j} graph. This is to be able to visualize the graph with many more nodes than NetworkX supports, and to make use of its Node2Vec feature.  
 

We used Neo4J to calculate the average, minimum, and maximum sentiment per review using the Query function.
The query retrieves all word/amenity nodes that are connected to the review. Then it finds all the relations between these nodes. If the sentiment of the relation is not NULL or 0, it is included for the sake of calculating the average sentiment. If there is no average sentiment to calculate, the average sentiment will simply be set to 0 by default. This average sentiment value is what we also use as a feature for the model. Of all these relations, the lowest and highest sentiment numbers are also stored on the node as features for the model. These values are also set to 0 by default. 
 
After this, we use the Neo4J Graph Data Science library to generate Node2Vec embeddings for our graph. We pick the review as the node to generate the embeddings on. Once that is done it will be stored on the nodes inside Neo4J. We then use a query to get a table of all nodes and their properties. This is downloaded as a json and then converted into a csv file using one of our scripts. The embeddings are then ready to use in addition to the average sentiment. Using these queries we can also store the lowest and highest sentiment relationships that are closely connected to the review if we want more features.
\subsubsection{Model Selection}
Once we have our features and embeddings we try to select the right model. Just as for our base model, we select Random Forest, Logistic Regression, and Multi-Layer Perceptron. As well as a dummy classifier.
\subsection{LLM-Based Model}
Our LLM-based model works by asking the LLM to predict the ratings of a set of reviews, based on the initial "training set" it is given, example-based learning \cite{ren2025malei,romero2025manchester}. So we first give it 200, or 2000 reviews with their corresponding ratings, and then based on those it has to predict the ratings for the other 9800, or 8000 reviews we give it next. To accomplish this task the LLM is allowed to write its own code.



\section{Experiments and Evaluations}
\label{experiments}

\subsection{Baseline Results}
For our baseline, we compared the three different classic NLP representation techniques using four different machine learning models to make the predictions with. The results can be seen in Table \ref{tab:model_comparison}. 


\begin{table}[t]
\centering
\small
\begin{tabular}{|l|c|c|c|c|}
\hline
\textbf{Model} & \textbf{Accuracy} & \textbf{MAE} & \textbf{RMSE} & \textbf{Cohen's Kappa} \\
\hline
\multicolumn{5}{|c|}{\textbf{Word2Vec}} \\
\hline
RF   & 0.57 & 0.58 & 0.98 & 0.30 \\
\textbf{LR}  & \textbf{0.60} & \textbf{0.50} & \textbf{0.79} & \textbf{0.38} \\
MLP  & 0.53 & 0.63 & 1.06 & 0.28 \\
MF   & 0.48 & 0.90 & 2.04 & 0.00 \\
\hline
\multicolumn{5}{|c|}{\textbf{Bag of Words}} \\
\hline
RF   & 0.54 & 0.73 & 1.54 & 0.17 \\
\textbf{LR}  & \textbf{0.60} & \textbf{0.49} & \textbf{0.69} & \textbf{0.38} \\
MLP  & 0.59 & 0.50 & 0.74 & 0.37 \\
MF   & 0.48 & 0.90 & 2.04 & 0.00 \\
\hline
\multicolumn{5}{|c|}{\textbf{TF-IDF}} \\
\hline
RF   & 0.53 & 0.76 & 1.60 & 0.15 \\
LR   & 0.60 & 0.51 & 0.80 & 0.35 \\
\textbf{MLP} & \textbf{0.59} & \textbf{0.50} & \textbf{0.72} & \textbf{0.36} \\
MF   & 0.48 & 0.90 & 2.04 & 0.00 \\
\hline
\end{tabular}
\caption{Evaluation results of machine learning models across different representation techniques. Values are rounded to two decimal places. "RF" = Random Forest, "LR" = Logistic Regression, "MLP" = Neural Network (Multi-Layer Perceptron), "MF" = Dummy (Most Frequent). Best values per metric are in bold.}
\label{tab:model_comparison}
\end{table}

\subsection{ReviewGraph Results}
For the following experiments, we exclusively use the Node2Vec embedding method, as implemented in the Neo4j Graph Data Science Library \cite{neo4j}. The model was configured with the following parameters: 1 iteration, embedding dimension of 10, walk length of 80, return factor of 1, in-out factor of 1, and a window size of 10. The embeddings were generated for nodes with the label \texttt{review}, using relationships of type \texttt{Any} and orientation \texttt{Any}. No relationship weight property was applied in this setup. These values may change if mentioned.
\subsubsection{Sampling Results}
Because 5-star ratings are significantly overrepresented in our dataset, leading to strong class imbalance, it was important to first evaluate which sampling strategy would yield the best results. We compared three approaches: no sampling, oversampling, and undersampling. Given the imbalance, our hypopaper was that oversampling would mitigate bias toward the majority class. While it might slightly reduce accuracy, we expected it to improve more meaningful metrics such as Root Mean Squared Error (RMSE) and Cohen’s Kappa. The basic model used is the Node2Vec embedding method with 10 embedding features, and that is what the 10 in the name "ReviewGraph-Node2Vec-10" stands for. These results used only the average sentiment of the review as additional feature. The classifier model used was just the Random Forest Classifier.


\begin{table}[t]
\centering
\begin{tabular}{|l|c|c|c|c|}
\hline
\textbf{Model Name} & \textbf{Acc.} & \textbf{MAE} & \textbf{RMSE} & \textbf{ Kappa} \\
\hline
RG-Node2Vec-10 & \textbf{0.49} & \textbf{0.81} & 1.69 & 0.03 \\
RG-Node2Vec-10-oversampling & 0.44 & 0.84 & \textbf{1.61} & \textbf{0.06} \\
RG-Node2Vec-10-undersampling & 0.27 & 1.41 & 3.45 & 0.05 \\
\hline
\end{tabular}
\caption{Performance of RG-Node2Vec-10 model with different sampling strategies. Values are rounded to two decimal places. "RG" = ReviewGraph. 
}
\label{tab:reviewgraph_sampling}
\end{table}

As shown in Table~\ref{tab:reviewgraph_sampling}, undersampling performed significantly worse across all metrics, likely due to the loss of valuable data. While the model without sampling achieved the highest accuracy and lowest MAE, it underperformed on RMSE and Cohen’s Kappa, which are more robust indicators of performance in imbalanced settings. Oversampling, as hypothesized, improved both RMSE and Kappa, suggesting more reliable predictions across the full range of rating classes. Based on these results, oversampling will be used as the default sampling strategy in all subsequent experiments.  
 
To illustrate the difference in sampling techniques we have the following figures:

\begin{figure}[t]
\centering
\begin{subfigure}[b]{0.4\textwidth}
    \includegraphics[width=\textwidth]{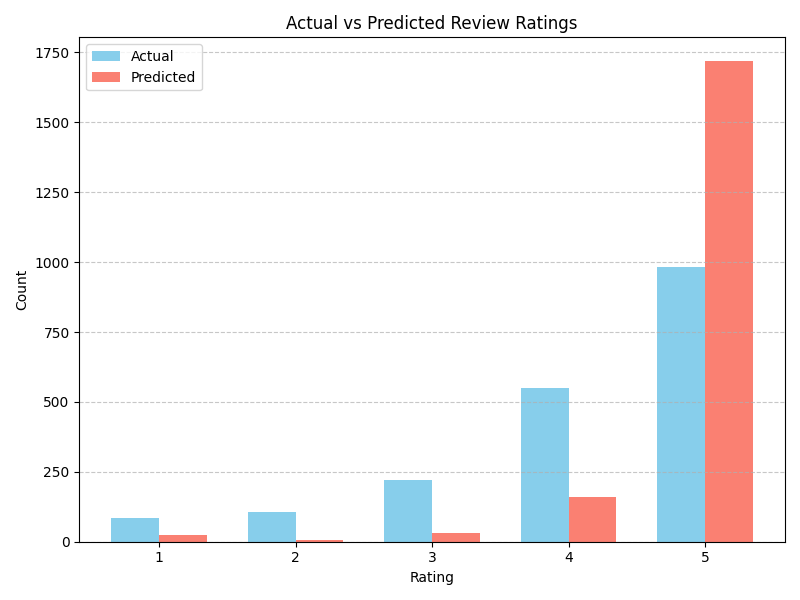}
    \caption{No Sampling}
    \label{fig:node2vec_nosampling}
\end{subfigure}
\hfill
\begin{subfigure}[b]{0.4\textwidth}
    \includegraphics[width=\textwidth]{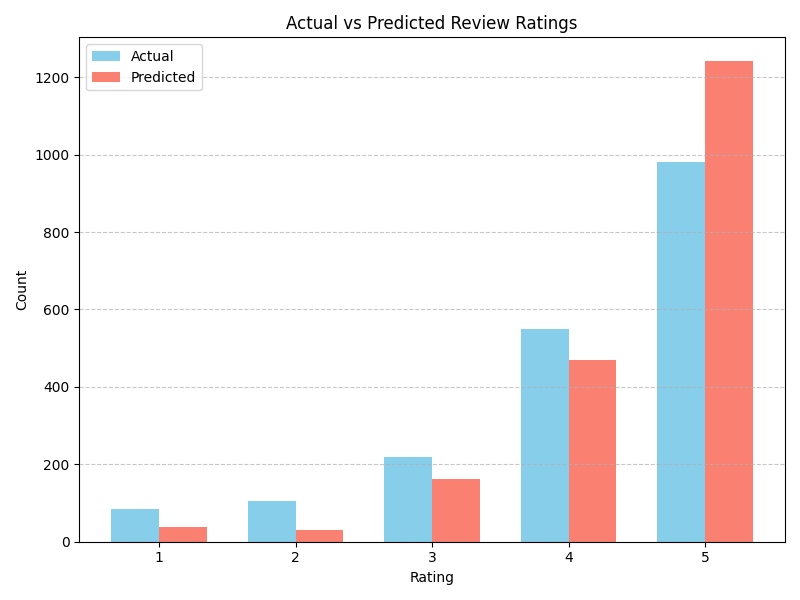}
    \caption{Oversampling}
    \label{fig:node2vec_oversampling}
\end{subfigure}
\hfill
\begin{subfigure}[b]{0.4\textwidth}
    \includegraphics[width=\textwidth]{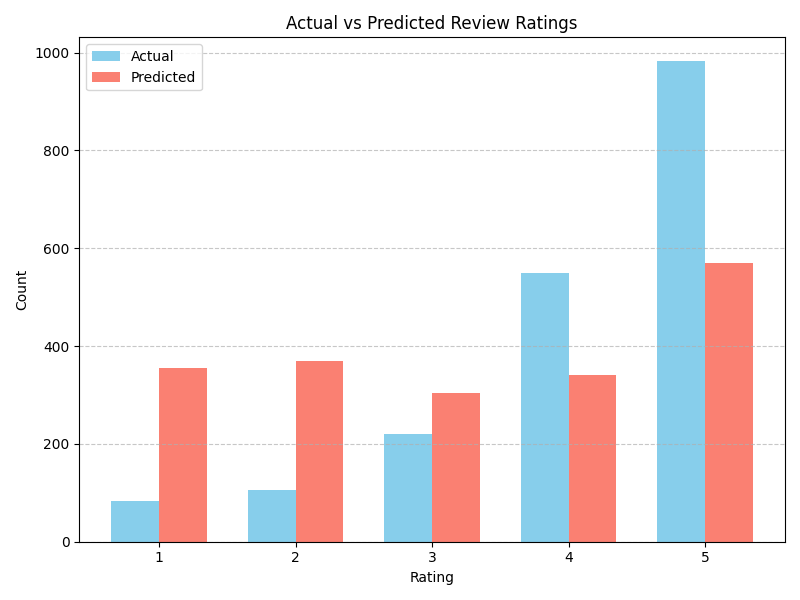}
    \caption{Undersampling}
    \label{fig:node2vec_undersampling}
\end{subfigure}
\caption{Prediction score distributions for Node2Vec-10 with different sampling strategies.}
\label{fig:node2vec_sampling_comparison}
\end{figure}

\subsubsection{Embedding Dimensions}
Next, we investigated the impact of embedding dimensionality on model performance. Using the oversampling strategy and the same Random Forest classifier as in previous experiments, we tested several different embedding sizes. Since Word2Vec typically uses 100 embedding dimensions by default, we used this as the upper limit for our experiments. The results are shown in Table~\ref{tab:node2vec_embedding_comparison}.

\begin{table}[t]
\centering
\begin{tabular}{|l|c|c|c|c|}
\hline
\textbf{Model Name} & \textbf{Acc.} & \textbf{MAE} & \textbf{RMSE} & \textbf{ Kappa} \\
\hline
RG-Node2Vec-5-oversampling & 0.42 & 0.85 & \textbf{1.58} & \textbf{0.07} \\
RG-Node2Vec-10-oversampling & 0.44 & 0.84 & 1.61 & 0.06 \\
RG-Node2Vec-25-oversampling & 0.46 & \textbf{0.81} & 1.60 & 0.05 \\
RG-Node2Vec-100-oversampling & \textbf{0.47} & 0.82 & 1.65 & 0.03 \\
\hline
\end{tabular}
\caption{Comparison of RG-Node2Vec model with different embedding dimensions (using oversampling). Values are rounded to two decimal places. 
}
\label{tab:node2vec_embedding_comparison}
\end{table}

Interestingly, the smallest embedding size of 5 produced the best RMSE and Cohen’s Kappa values, suggesting more consistent and robust predictions. As the number of dimensions increased, RMSE and Kappa generally worsened, potentially indicating overfitting or noise amplification in higher-dimensional spaces. Although accuracy increased with more dimensions, peaking at 100, it is likely this reflects a growing bias toward the dominant class (5-star ratings), rather than genuine improvements in prediction quality. This trend emphasizes the importance of evaluating models with metrics beyond accuracy, particularly in imbalanced settings like this one.

\subsubsection{Model Selection}
We are also interested in which classifier model to use to predict the ratings. We decided to compare some common classifiers, Random Forest, Logistic Regression (Max Entropy Loss), and an MLP Neural Network Classifier. Additionally we tested it with a most frequent dummy, with no sampling in that case, as this would not work when oversampling is used. Scaling is used in the case of the neural network and logistic regression.

\begin{table}[t]
\centering
\begin{tabular}{|l|c|c|c|c|c|c|}
\hline
\textbf{Model} & \textbf{Dim} & \textbf{Sampling} & \textbf{Acc.} & \textbf{MAE} & \textbf{RMSE} & \textbf{ Kappa} \\
\hline
\textbf{RF} & 100 & Oversampling & \textbf{0.47} & \textbf{0.81} & \textbf{1.64} & \textbf{0.04} \\
LR & 100 & Oversampling & 0.20 & 1.73 & 4.61 & 0.02 \\
MLP & 100 & Oversampling & 0.18 & 1.33 & 2.49 & -0.00 \\
\hline
\textbf{RF} & 5 & Oversampling & \textbf{0.41} & \textbf{0.87} & \textbf{1.64} & \textbf{0.06} \\
LR & 5 & Oversampling & 0.23 & 1.62 & 4.37 & 0.03 \\
MLP & 5 & Oversampling & 0.17 & 1.85 & 5.06 & 0.00 \\
\hline
\textbf{RF} & 5 & No Sampling & 0.47 & \textbf{0.83} & \textbf{1.71} & 0.02 \\
LR & 5 & No Sampling & 0.31 & 1.53 & 4.30 & \textbf{0.06} \\
MLP & 5 & No Sampling & 0.23 & 1.37 & 3.26 & -0.01 \\
Dummy & 5 & No Sampling & \textbf{0.51} & 0.85 & 1.92 & 0.00 \\
\hline
\textbf{RF} & 100 & No Sampling & 0.50 & \textbf{0.85} & \textbf{1.91} & -0.01 \\
LR & 100 & No Sampling & 0.33 & 1.48 & 4.28 & \textbf{0.08} \\
MLP & 100 & No Sampling & 0.27 & 1.21 & 2.63 & 0.00 \\
Dummy & 100 & No Sampling & \textbf{0.51} & 0.85 & 1.92 & 0.00 \\
\hline
\end{tabular}
\caption{Performance comparison of different classifiers using ReviewGraph-Node2Vec embeddings, varying embedding dimensions and sampling strategies. 
}
\label{tab:merged_node2vec_model_comparison}
\end{table}

As shown in Table~\ref{tab:merged_node2vec_model_comparison}, the Random Forest classifier consistently outperforms the other models across most configurations, especially when combined with oversampling. It achieves the best overall accuracy, MAE, and RMSE scores in both the 5- and 100-dimensional embedding setups. Logistic Regression performs reasonably well in terms of Cohen’s Kappa, especially under the no sampling condition, indicating that it may be slightly more sensitive to class distribution. However, its overall performance in terms of error metrics is considerably weaker. The Neural Network model shows poor performance across all configurations, likely due to overfitting and insufficient tuning or data volume. The dummy classifier performs surprisingly well in accuracy under no sampling, but this result is misleading, as it simply predicts the majority class (5 stars) and fails to generalize. Based on these results, we select the Random Forest model as the classifier for subsequent experiments.
\subsubsection{Complex Sentiment Analysis}
Lastly, we are interested in a more nuanced and complex sentiment analysis, moving beyond average sentiment values to consider the full range of emotional tone expressed in each review. Specifically, we examine the lowest and highest sentiment ratings associated with semantic relationships between word/amenity nodes connected to the same review.

Each review is connected to entities such as words or amenities via a \texttt{CONTAINS} relationship. By analyzing the sentiment values on the edges between these connected nodes we capture not only the overall sentiment but also the extremes within the sentiment distribution.

The minimum sentiment score highlights the most negative relationship sentiment mentioned in the review, while the maximum score captures the most positive sentiment. This approach helps identify reviews that may express a mix of strong praise and criticism, which would otherwise be obscured in average-based analysis.

These min/max sentiment scores are computed and stored for each review using a Neo4J query, allowing downstream models to leverage this richer representation of sentiment variance. We believe this adds valuable interpretability and robustness when modeling customer feedback and satisfaction levels.

\begin{table}[t]
\centering
\begin{tabular}{|l|l|c|c|c|c|}
\hline
\textbf{Model} & \textbf{Sampling} & \textbf{Acc.} & \textbf{MAE} & \textbf{RMSE} & \textbf{Cohen’s Kappa} \\
\hline
\textbf{RF} & No Sampling & \textbf{0.56} & \textbf{0.60} & \textbf{1.02} & \textbf{0.26} \\
LR & No Sampling & 0.36 & 1.31 & 3.50 & 0.17 \\
MLP & No Sampling & 0.46 & 1.08 & 2.83 & 0.09 \\
Dummy  & No Sampling & 0.51 & 0.85 & 1.92 & 0.00 \\
\hline
\textbf{RF} & Oversampling & \textbf{0.54} & \textbf{0.63} & \textbf{1.09} & \textbf{0.28} \\
LR & Oversampling & 0.43 & 1.13 & 3.01 & 0.21 \\
MLP & Oversampling & 0.50 & 0.82 & 1.80 & 0.09 \\
Dummy  & Oversampling & 0.04 & 3.15 & 11.15 & 0.00 \\
\hline
\end{tabular}
\caption{Performance metrics of various ML models with min and max sentiment scores, with and without oversampling. 
}
\label{tab:ml_oversampling_results}
\end{table}

Compared to the previous results, the performance of this model shows notable improvements across most evaluation metrics. In particular, the Cohen’s Kappa score benefits significantly from the incorporation of richer sentiment features, indicating a stronger agreement between predicted and actual ratings beyond chance. This suggests that the use of more expressive sentiment representations, such as minimum and maximum relational sentiment, enhances the model’s ability to capture nuanced review patterns.

\subsection{LLM Results}
To evaluate whether an LLM can predict ratings better than our baseline or ReviewGraph model, we made use of GPT-4o. We first gave it a randomly sampled set of 200, and 2000 reviews of our total 10,000 dataset. These reviews were given alongside the ratings. We also \textit{allowed the model to write code to do the predictions} for it. It decided to make a TF-IDF vectorizer plus logistic regression classifier, similar to one of our baselines. The evaluation results can be seen in table ~\ref{tab:model-comparison-llm}.

\begin{table}[t]
\centering
\begin{tabular}{|l|l|c|c|c|c|}
\hline
\textbf{Model} & \textbf{Sampling Size} & \textbf{Acc.} & \textbf{MAE} & \textbf{RMSE} & \textbf{ Kappa} \\
\hline
LLM Model & n=200 & 0.52 & 0.81 & 1.77 & 0.04 \\
LLM Model & n=2000 & \textbf{0.59} & \textbf{0.58} & \textbf{1.06} & \textbf{0.28} \\
\hline
\end{tabular}
\caption{Performance metrics of the LLM-based rating prediction model using different training set sizes. Values are rounded to two decimal places; bold indicates the best-performing score per metric.}
\label{tab:model-comparison-llm}
\end{table}

\subsection{Ablation studies}
To test how much different features actually impact our ReviewGraph model, we tested it without various parts of the model. Firstly we wanted to see what happens if we only make use of the average, min, and max sentiment. Without the Node2Vec embeddings. Results can be seen in Table ~\ref{tab:maxmin_full_results_no_node2vec}

\begin{table}[t]
\centering
\begin{tabular}{|l|l|c|c|c|c|}
\hline
\textbf{Model} & \textbf{Sampling} & \textbf{Acc.} & \textbf{MAE} & \textbf{RMSE} & \textbf{Kappa} \\
\hline
RF & No Sampling & \textbf{0.48} & \textbf{0.77} & \textbf{1.47} & \textbf{0.14} \\
LR & No Sampling & 0.43 & 1.19 & 3.19 & 0.13 \\
MLP & No Sampling & 0.40 & 1.38 & 4.02 & 0.09 \\
Dummy & No Sampling & 0.51 & 0.85 & 1.92 & 0.00 \\
\hline
RF & Oversampling & 0.39 & \textbf{1.02} & \textbf{2.26} & 0.12 \\
LR & Oversampling & \textbf{0.44} & 1.12 & 2.92 & \textbf{0.14} \\
MLP & Oversampling & 0.37 & 1.43 & 4.17 & 0.11 \\
Dummy & Oversampling & 0.04 & 3.15 & 11.15 & 0.00 \\
\hline
\end{tabular}
\caption{Performance metrics of the ReviewGraph model without Node2Vec embeddings.}
\label{tab:maxmin_full_results_no_node2vec}
\end{table}

It performs clearly worse than our model in the Complex Sentiment Analysis section, that does make use of the Node2Vec embeddings. 
 
Next we were curious how the model would perform if it only uses the Node2Vec embeddings. In this case we are using the 5 dimensional embedding as that was the best performing Node2Vec dimension.

\begin{table}[t]
\centering
\begin{tabular}{|l|l|c|c|c|c|}
\hline
\textbf{Model} & \textbf{Sampling} & \textbf{Acc.} & \textbf{MAE} & \textbf{RMSE} & \textbf{ Kappa} \\
\hline
RF & No Sampling & \textbf{0.54} & \textbf{0.65} & \textbf{1.17} & \textbf{0.24} \\
LR & No Sampling & 0.34 & 1.25 & 3.15 & 0.15 \\
MLP & No Sampling & 0.46 & 1.11 & 2.96 & 0.05 \\
Dummy & No Sampling & 0.51 & 0.85 & 1.92 & 0.00 \\
\hline
RF & Oversampling & \textbf{0.50} & \textbf{0.71} & \textbf{1.27} & \textbf{0.23} \\
LR & Oversampling & 0.38 & 1.17 & 2.99 & 0.16 \\
MLP & Oversampling & 0.30 & 1.10 & 2.34 & 0.05 \\
Dummy & Oversampling & 0.04 & 3.15 & 11.15 & 0.00 \\
\hline
\end{tabular}
\caption{Performance metrics of ML models using only 5 dimensional Node2Vec embedding features, with and without oversampling.}
\label{tab:node2vec5_combined_results}
\end{table}

Interestingly, the models using only the 5-dimensional Node2Vec embeddings without any sentiment or additional features performed surprisingly well. In particular, the Random Forest model achieved the highest accuracy and lowest error metrics within this configuration, with a Cohen’s Kappa of 0.2393, comparable to our results in the Complex Sentiment Analysis section. This suggests that the structural information encoded by the Node2Vec embeddings alone captures a substantial amount of predictive signal regarding review ratings. It also performed way better than our model that was using 5 dimensions of Node2Vec embeddings and only average sentiment in addition. However, when using the embeddings, in addition to average, max, and min sentiment, the scores are higher than just the Node2Vec embeddings.

\subsection{Discussion}
The ReviewGraph model with the highest Cohen's Kappa score ended up using 5 embedding dimensions, oversampling, and made use of the complex sentiment features (avg, min, max). The random forest model also consistently performed the best on ReviewGraph results, while it did not on the baseline models. We believe this is because our model is lower dimensional, especially when using only 5 node embeddings. This means that it is easier for random forests to split trees on.

Between the best results of the baseline, ReviewGraph model, and the LLM model there was not a substantial difference. Our current best approach has a similar Cohen's Kappa to the LLM model, while having a slightly lower accuracy score. The Cohen's Kappa is however substantially higher for the best baselines, while the accuracy scores are not much higher. The reason for this could be because we used a fairly crude triple extraction method. It generated a lot of low quality triples that had to be disregarded. Additionally, many of the triples or nodes only show up once and thus are not useful data for the machine learning algorithm. We did not drop these nodes before generating the Node2Vec embeddings. 
 
Additionally, while we calculate the sentiments for all the triples our model does currently not use all of those sentiments to make predictions. Our results show that just adding the minimum and maximum sentiments increases the accuracy, and especially Cohen's Kappa very substantially. This means that a lot of context is lost.

\section{Visualising ReviewGraph}
The graph structure of ReviewGraph demonstrates significant potential as a valuable data visualization tool for understanding customer feedback.
Currently, we developed several minor features for this graph visualization, which lets the user sort the graph to show data they are specifically interested in. More features can easily be added in the future.
Some of the current features are:
\begin{itemize}
    \item Filtering reviews by a specified score or score range (e.g., 2–4).
    \item Highlighting reviews, hotels, or keywords based on highest or lowest sentiment scores.
    \item Enabling users to inspect low-sentiment reviews by displaying their associated nodes, providing a high-level overview of the review’s content.
\end{itemize}

\begin{figure}[t]
    \centering
    \includegraphics[width=1\linewidth]{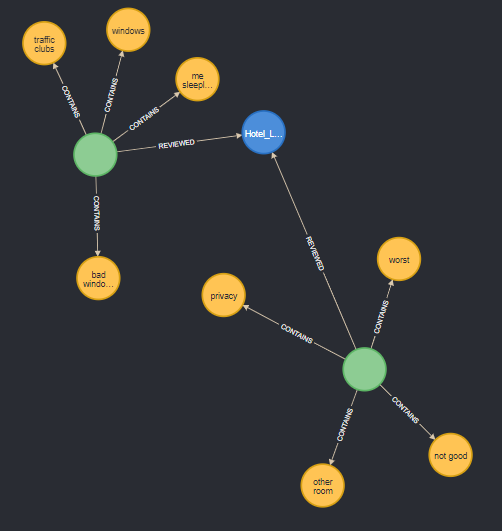}
    \caption{Example of a hotel and two negative sentiment reviews connected to it. The blue circle is the hotel, the green ones are the reviews, and the yellow ones are subjects/objects extracted from the review text. }
    \label{fig:system-visualisation}
\end{figure}

\begin{figure}[t]
    \centering
    \includegraphics[width=1\linewidth]{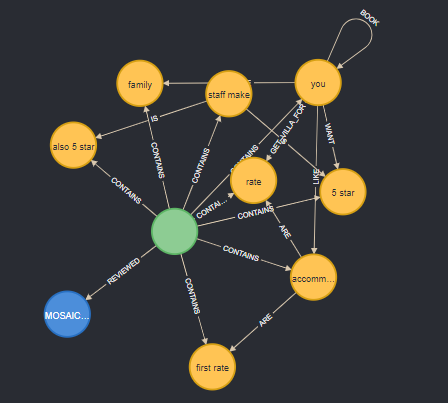}
    \caption{Example of review and all the nodes connected to it as well as the relationships between those nodes connected to it. The blue one is the hotel, the green one the review, and the yellow ones are subjects/object extracted from the review text. }
    \label{fig:system-visualisation-review}
\end{figure}

Figure \ref{fig:system-visualisation} and Figure \ref{fig:system-visualisation-review} are examples of how the visualisation currently works. Every node is interactable and it shows all the nodes it is connected to when clicked on.

\section{Conclusions}\label{conclusions}

To explore the potential of KG embedding on relation represeantion and prediction, we investigated the ReviewGraph methodology that we proposed for review prediction task. The KG-embedding based model performed very compatitively to the traditional baseline models that we implemented. We believe there is much space to further improve ReviewGraph such as by deploying differnt embedding models, which opens up a new oppotunity for future research. To the best of our knowledge, we are the first to propose ReviewGraph model with KG-embeddings integrated to the applicaiton task on review predictions.


We also believe that the graph structure of our model opens up new avenues, in terms of potentially visualizing the model for users, making use of GraphRAG to have LLMs summarise the contents of the graph and by extension the reviews, and suggesting actionable insights by doing a prediction before and after certain nodes are removed. Visualisation graphs, key codes, and 10-fold cross validations can been found at our open source website \url{https://github.com/aaronlifenghan/ReviewGraph}
 

\subsection{Limitations and Further Research}
Node2Vec is a relatively simple algorithm that only looks at the neighboring nodes and not the relations, or graph structure between the neighboring nodes. It also can only be used on one specific graph, if even one node is added the Node2Vec algorithm has to be re-trained. For this reason, we believe investigating more advanced algorithms such as GraphSAGE or other types of Graph Neural Networks to experiment on this knowledge graph has potential. 
 
Additionally, different graph structures and preprocessing steps could be tried. It could also be possible that different classifier models would be more effective on this use case. 
 
Initially, we also investigated whether LLMs could be used for the task of triple extraction, but opted not to because of computational limits. We believe using a fine-tuned LLM for triple extraction has more potential than using StanfordOpenIE and could have led to more effective feature extraction, and by extension, more accurate rating prediction. Many of the triples in our dataset were low quality, often way too long in length, and we believe a fine-tuned LLM would not have this issue as much. 
 
Additionally, we believe there is potential in investigating whether the graph representation of reviews is an effective tool for hoteliers to be able to quickly see what reviewers are generally talking about. And whether they are speaking about those topics negatively or positively. Doing tests with human evaluators to see if this visualisation is effective is a potential avenue of research.  
 
We also see potential for utilizing GraphRAG to quickly summarize the review data using an LLM. And to also have the LLM suggest actionable insights based on a prediction before and after a certain node is removed. For example, have the model predict what would happen if the bathrooms were cleaner, by removing the corresponding negative cleanliness-related nodes connected to bathroom nodes.

\bibliography{nodalida2023}
\bibliographystyle{IEEEtran}

\newpage

\appendix
\section*{Sample of Extracted Triples}
\label{appendix:triples}

To illustrate the output of the triple extraction process used in this study, we include a representative sample of 20 triples extracted from different user reviews. These triples reflect a range of sentiment scores (positive, negative, and neutral) and capture various aspects of the reviews. Each triple is structured as $\langle$ \textit{Subject, Relation, Object} $\rangle$, with an associated sentiment score. Inside our knowledge graph we remove all triples with relationships or nodes that have a text length of longer than 14 characters.

\begin{table*}[t]
\centering
\begin{tabular}{|c|p{3.2cm}|p{2.8cm}|p{3.2cm}|c|}
\hline
\textbf{Review ID} & \textbf{Subject} & \textbf{Relation} & \textbf{Object} & \textbf{Sentiment} \\
\hline
144   & Great pool & is in & wonderful spot by beach & 0.83 \\
3798  & bed & was comfortable with & excellent linen & 0.79 \\
992   & positive & were many small issues with & room for improvement & 0.77 \\
6254  & loft & best feature of was & bathroom & 0.64 \\
276   & Breakfast & was outstanding for & British fryup & 0.61 \\
1299  & we & were & most impressed & 0.53 \\
2185  & Our 40th school reunion weekend & was & help & 0.40 \\
1513  & hotel & is well located in & historical center & 0.27 \\
4683  & Hotel & was accommodating to & our group & 0.00 \\
5108  & We & brought along & our 8yearold & 0.00 \\
721   & We & had & 5night stay & 0.00 \\
6958  & Ravenna & was & crowded & 0.00 \\
3744  & same & can & can said of bathroom & 0.00 \\
8153  & This & is & our 4th year staying here & 0.00 \\
5882  & manager & moved with & only minor change fee & 0.00 \\
9326  & it & is & too much trouble & -0.40 \\
6644  & Poor excuse & is in & need & -0.42 \\
5672  & water temperature & keeps & fluctuating dangerously & -0.46 \\
8623  & check & is & wrong & -0.48 \\
4844  & My complaint & was & very poor wireless internet service & -0.68 \\
\hline
\end{tabular}
\caption{Sample of Extracted Triples from Review Data}
\label{tab:sample_triples}
\end{table*}

\section*{Key Implementation Code}
This section includes selected code snippets that are central to the implementation of our model. The complete source code can be found in our GitHub repository.

\begin{center}
\url{https://github.com/aaronlifenghan/ReviewGraph}
\end{center}

\section*{Complete Tables of Results}
\subsection{Train Test Split Results}

\begin{table*}[t!]
\centering
\scriptsize
\begin{tabular}{|l|l|c|l|c|c|c|c|}
\hline
\textbf{Classifier Model} & \textbf{Type} & \textbf{Dim} & \textbf{Sampling} & \textbf{Accuracy} & \textbf{MAE} & \textbf{RMSE} & \textbf{Cohen’s Kappa} \\
\hline
Random Forest & Node2Vec & 100 & Oversampling & 0.4709 & 0.8135 & 1.6409 & 0.0352 \\
Logistic Regression & Node2Vec & 100 & Oversampling & 0.1968 & 1.7331 & 4.6121 & 0.0180 \\
Neural Network (MLP) & Node2Vec & 100 & Oversampling & 0.1757 & 1.3282 & 2.4874 & -0.0025 \\
Random Forest & Node2Vec & 5 & Oversampling & 0.4060 & 0.8722 & 1.6388 & 0.0574 \\
Logistic Regression & Node2Vec & 5 & Oversampling & 0.2334 & 1.6249 & 4.3699 & 0.0307 \\
Neural Network (MLP) & Node2Vec & 5 & Oversampling & 0.1664 & 1.8537 & 5.0644 & 0.0043 \\
Random Forest & Node2Vec & 5 & No Sampling & 0.4673 & 0.8315 & 1.7105 & 0.0188 \\
Logistic Regression & Node2Vec & 5 & No Sampling & 0.3060 & 1.5281 & 4.2998 & 0.0562 \\
Neural Network (MLP) & Node2Vec & 5 & No Sampling & 0.2257 & 1.3735 & 3.2643 & -0.0070 \\
Dummy (Most Frequent) & Node2Vec & 5 & No Sampling & 0.5059 & 0.8465 & 1.9201 & 0.0000 \\
Random Forest & Node2Vec & 100 & No Sampling & 0.4951 & 0.8542 & 1.9134 & -0.0076 \\
Logistic Regression & Node2Vec & 100 & No Sampling & 0.3344 & 1.4776 & 4.2751 & 0.0751 \\
Neural Network (MLP) & Node2Vec & 100 & No Sampling & 0.2700 & 1.2081 & 2.6321 & 0.0049 \\
Dummy (Most Frequent) & Node2Vec & 100 & No Sampling & 0.5059 & 0.8465 & 1.9201 & 0.0000 \\
Random Forest & N2V+Sentiment(min/max) & 5 & No Sampling & 0.5590 & 0.5992 & 1.0185 & 0.2607 \\
Logistic Regression & N2V+Sentiment(min/max) & 5 & No Sampling & 0.3632 & 1.3060 & 3.5018 & 0.1688 \\
Neural Network (MLP) & N2V+Sentiment(min/max) & 5 & No Sampling & 0.4626 & 1.0752 & 2.8300 & 0.0937 \\
Dummy (Most Frequent) & N2V+Sentiment(min/max) & 5 & No Sampling & 0.5059 & 0.8465 & 1.9201 & 0.0000 \\
Random Forest & N2V+Sentiment(min/max) & 5 & Oversampling & 0.5404 & 0.6301 & 1.0896 & 0.2829 \\
Logistic Regression & N2V+Sentiment(min/max) & 5 & Oversampling & 0.4250 & 1.1345 & 3.0057 & 0.2107 \\
Neural Network (MLP) & N2V+Sentiment(min/max) & 5 & Oversampling & 0.5039 & 0.8238 & 1.7965 & 0.0862 \\
Dummy (Most Frequent) & N2V+Sentiment(min/max) & 5 & Oversampling & 0.0433 & 3.1535 & 11.1484 & 0.0000 \\
Random Forest & Word2Vec & -- & -- & 0.5705 & 0.5775 & 0.9805 & 0.2988 \\
Logistic Regression & Word2Vec & -- & -- & \textbf{0.6035} & \textbf{0.5010} & \textbf{0.7920} & \textbf{0.3793} \\
Neural Network (MLP) & Word2Vec & -- & -- & 0.5275 & 0.6325 & 1.0555 & 0.2828 \\
Dummy (Most Frequent) & Word2Vec & -- & -- & 0.4765 & 0.9025 & 2.0405 & 0.0000 \\
Random Forest & Bag of Words & -- & -- & 0.5350 & 0.7335 & 1.5405 & 0.1704 \\
Logistic Regression & Bag of Words & -- & -- & 0.5960 & 0.4855 & 0.6875 & 0.3802 \\
Neural Network (MLP) & Bag of Words & -- & -- & 0.5920 & 0.5010 & 0.7360 & 0.3698 \\
Dummy (Most Frequent) & Bag of Words & -- & -- & 0.4765 & 0.9025 & 2.0405 & 0.0000 \\
Random Forest & TF-IDF & -- & -- & 0.5265 & 0.7550 & 1.6010 & 0.1500 \\
Logistic Regression & TF-IDF & -- & -- & 0.5955 & 0.5120 & 0.7990 & 0.3474 \\
Neural Network (MLP) & TF-IDF & -- & -- & 0.5880 & 0.5015 & 0.7165 & 0.3607 \\
Dummy (Most Frequent) & TF-IDF & -- & -- & 0.4765 & 0.9025 & 2.0405 & 0.0000 \\
LLM Model & LLM (n=200) & -- & -- & 0.5165 & 0.8050 & 1.7748 & 0.0427 \\
LLM Model & LLM (n=2000) & -- & -- & 0.5867 & 0.5795 & 1.0622 & 0.2839 \\
\hline
\end{tabular}
\caption{Performance comparison across all classifier models, feature types, sampling methods, and dimensions. Bold values indicate the best overall scores across all configurations.}
\label{tab:mega_model_comparison}
\end{table*}

\subsection{10-Fold Cross Validation Results}
\begin{table*}[t!]
\centering
\scriptsize
\begin{tabular}{|l|l|c|l|c|c|c|c|}
\hline
\textbf{Classifier Model} & \textbf{Type} & \textbf{Dim} & \textbf{Sampling} & \textbf{Accuracy} & \textbf{MAE} & \textbf{RMSE} & \textbf{Cohen’s Kappa} \\
\hline
Random Forest & Node2Vec & 5 & Oversampling & 0.4094 & 0.8766 & 1.6609 & 0.0558 \\
Logistic Regression & Node2Vec & 5 & Oversampling & 0.3575 & 1.4730 & 4.3033 & 0.0807 \\
Neural Network (MLP) & Node2Vec & 5 & Oversampling & 0.2765 & 1.3625 & 3.2776 & 0.0601 \\
Dummy (Most Frequent) & Node2Vec & 5 & Oversampling & 0.0406 & 3.1612 & 11.1686 & 0.0000 \\
Random Forest & Node2Vec & 5 & No Sampling & 0.4771 & 0.8121 & 1.6404 & 0.0400 \\
Logistic Regression & Node2Vec & 5 & No Sampling & 0.5085 & 0.8355 & 1.8657 & 0.0073 \\
Neural Network (MLP) & Node2Vec & 5 & No Sampling & 0.4917 & 0.8190 & 1.7327 & 0.0254 \\
Dummy (Most Frequent) & Node2Vec & 5 & No Sampling & 0.5080 & 0.8388 & 1.8789 & 0.0000 \\
Random Forest & N2V+Sentiment(min/max) & 5 & Oversampling & 0.5243 & 0.6379 & 1.0605 & 0.2615 \\
Logistic Regression & N2V+Sentiment(min/max) & 5 & Oversampling & 0.4913 & 0.7569 & 1.4742 & 0.2649 \\
Neural Network (MLP) & N2V+Sentiment(min/max) & 5 & Oversampling & 0.4543 & 0.8389 & 1.6516 & 0.2249 \\
Dummy (Most Frequent) & N2V+Sentiment(min/max) & 5 & Oversampling & 0.0406 & 3.1612 & 11.1686 & 0.0000 \\
Random Forest & Word2Vec & -- & -- & 0.5942 & 0.5365 & 0.8971 & 0.3213 \\
Logistic Regression & Word2Vec & -- & -- & 0.6286 & 0.4605 & 0.6981 & \textbf{0.4055} \\
MLP Classifier & Word2Vec & -- & -- & 0.5461 & 0.5874 & 0.9336 & 0.2957 \\
Dummy (Most Frequent) & Word2Vec & -- & -- & 0.5060 & 0.8380 & 1.8686 & 0.0000 \\
Random Forest & Bag of Words & -- & -- & 0.5612 & 0.6751 & 1.3739 & 0.1888 \\
Logistic Regression & Bag of Words & -- & -- & 0.6113 & 0.4608 & \textbf{0.6334} & 0.3855 \\
MLP Classifier & Bag of Words & -- & -- & 0.6056 & 0.4746 & 0.6694 & 0.3751 \\
Dummy (Most Frequent) & Bag of Words & -- & -- & 0.5060 & 0.8380 & 1.8686 & 0.0000 \\
Random Forest & TF-IDF & -- & -- & 0.5619 & 0.6839 & 1.4245 & 0.1848 \\
Logistic Regression & TF-IDF & -- & -- & \textbf{0.6367} & \textbf{0.4536} & 0.6942 & 0.4007 \\
MLP Classifier & TF-IDF & -- & -- & 0.5966 & 0.4817 & 0.6695 & 0.3591 \\
Dummy (Most Frequent) & TF-IDF & -- & -- & 0.5060 & 0.8380 & 1.8686 & 0.0000 \\
\hline
\end{tabular}
\caption{10-Fold Cross-Validation Performance comparison across all classifier models, feature types, sampling methods,
and dimensions. Bold values indicate the best overall scores across all configurations.}
\label{tab:tenfold_model_comparison}
\end{table*}

\end{document}